\documentclass[a4paper,twoside]{article}

\usepackage{epsfig}
\usepackage{subfigure}
\usepackage{calc}
\usepackage{amssymb}
\usepackage{amstext}
\usepackage{amsmath}
\usepackage{amsthm}
\usepackage{multicol}
\usepackage{pslatex}
\usepackage{natbib}
\let\bibhang\relax
\usepackage{apalike}
\usepackage{pdfpages}
\usepackage{SCITEPRESS}     % Please add other packages that you may need BEFORE the SCITEPRESS.sty package.

\begin{document}

\title{GAdaBoost: Accelerating Adaboost Feature Selection with Genetic Algorithms  }

\author{\authorname{Mai F. Tolba\sup{1}, Mohamed Moustafa\sup{1}}
\affiliation{\sup{1}Computer Science and Engineering Department, The American University In Cairo, Road 90, New Cairo, Cairo, Egypt}
\email{maitolba@aucegypt.edu, m.moustafa@aucegypt.edu}}


\abstract{Boosted cascade of simple features, by Viola and Jones, is one of the most famous object detection frameworks. However, it suffers from a lengthy training process. This is due to the vast features space and the exhaustive search nature of Adaboost. In this paper we propose GAdaboost: a Genetic Algorithm to accelerate the training procedure through natural feature selection. Specifically, we propose to limit Adaboost search within a subset of the huge feature space, while evolving this subset following a Genetic Algorithm. Experiments demonstrate that our proposed GAdaboost is up to 3.7 times faster than Adaboost. We also demonstrate that the price of this speedup is a mere decrease (3\%, 4\%) in detection accuracy when tested on FDDB benchmark face detection set, and Caltech Web Faces respectively.}

\onecolumn \maketitle \normalsize \vfill

\section{\uppercase{Introduction}}
\label{sec:INTRODUCTION}
\noindent Machine learning and training, require large feature sets, which can be time consuming to explore. With the advancement of this field the need for algorithms to decrease the training time arose. Genetic Algorithms (GA) has proven its strength in solving problems like the aforementioned one, especially those concerned with exploring large search spaces, and providing acceptable results in a significantly less amount of time than the brute force manner. Many research has explored the use of GA in time consuming tasks like feature selection, which aims to choose a representative small subset of features from the whole set of features
\citep{Xue2015}.

Object detection lies in the set of machine learning techniques that require a huge search space for training, thus their training is time consuming. Object detection is concerned with detecting whether an object is present in a given image, and where it lies in this image. It has many applications including, but not limited to, face detectors in all modern state of the art cameras, automotive safety, video indexing, image classification, surveillance, and content-based image retrieval \citep{Lillywhite2013}.

Much research has been put into this area, due to its complex nature as detection is hard to achieve in different light conditions, occlusion and the angle in which the object appears in the image \citep{Lienhart2002,Lillywhite2013,Viola2001}. Researchers has been trying to implement efficient high speed detectors that work in real time and has a high percentage of accuracy. Though the Viola-Jones detector has reached an impressive detection speed it still consumes a lot of time in training. Viola-Jones uses Adaboost, a type of boosting algorithms, to select and combine weak classifiers to form a strong one. Adaboost is simple and adaptive \citep{Dezhen2008}, yet it operates in a brute force manner, passing by all the set of features multiple times. This can be very time consuming, as the search space consists of a set of more than 162,000 features for a 24X24 image. 

The main contribution of this paper is increasing the speed of training of the Viola-Jones face detector by implementing a hybrid approach that combines the use of boosting and GAs. This will find the best features efficiently instead of going through all of them in a brute force manner, which will decrease the training time. The Paper is organized as follows: section 2 discusses some of the related previous work. Section 3 explains the proposed method. Section 4 provides some experiments and their results. Section 5 concludes the paper and discuss future enhancements.

\section{\uppercase{RELATED WORK}}
\label{sec:RELATED WORK}

\noindent Genetic Algorithms (GA) are optimizing procedures that are devised from the biological mechanism of reproduction, and evolutionary science \citep{Sun2004}. GA continues to prove itself successful in many fields including object detection.
There are other optimization methods that serve well; however, from some experiments, GA has proven to perform better in solving problems. This might be due to the advantages of GA and that they are probabilistic and not deterministic, and have the ability to be better at avoiding to be stuck at a local maxima and are parallelizable.
 \citet{Ferri1994403}, compared GA against sequential search and their results clearly show that GA performs better. Their work highlighted the point of strength of GA which is the ability to perform the search in a near optimal region due to the inherited randomizations used in the search. \citet{Tabassum2014}, said that “It was proved that genetic algorithms are the most powerful unbiased optimization techniques for sampling a large solution space. After implementing the Knapsack problem and image optimization, they concluded in their paper that GA are the best application to solve various common problems and that they are suitable for solving high complexity problems like the combinatorial optimizations. \citet{Sun2004} provided another proof to the strength of GA, when they used it to select the best eigenvectors. In their work GA was used to solve the problem of selecting the best feature set. They compared their results with other techniques and proved to provide better accuracy with less number of features. \citet{Lillywhite2013} used Genetic algorithms in constructing features which was used by Adaboost to train a classifier. They tested their approach against previously published papers and used the same dataset for comparison. Their technique proved to be significantly more accurate than most of the previous work they compared against. Some researchers used GA in feature selection. Feature selection methods can be divided into 3 main categories: wrappers, filters and embedded methods. Filters are a form of preselecting each feature on its own without considering the previous predictor. Wrappers are methods to score the predictive power of a subset of features by using a machine learning technique as a black box, while the embedded are methods that integrate classification and learning into a single process
 \citep{Chouaib2008,Xue2015}
. In Their work, \citet{Chouaib2008} aimed to find the set of the most representative features using GAs, in order to decrease the detection time. Their results showed that for the majority of descriptors their feature set was significantly reduced up to 75\% of the original set in two class problems. \citet{Dezhen2008} provided a post optimization technique to avoid the redundancy of classifiers. By doing so, they managed to increase the speed of classification by 110\% due to reducing the number of features to 55\% of the original set. Since this is a post optimization process, it can be considered as an added part to the training process, which will be an overhead on the training time. \citet{Xue2015} provided a survey on the use of evolutionary computing in feature selection. In their work they surveyed more than 40 papers which use GA in feature selection.

Object detection is a main area of research in computer vision. It falls under the type of problems that suffer from a time consuming training process, due to the huge search space involved. \citet{Viola2001} devised a new face detector using Haar features, since features provide a set of comprehensive information that can be learned by machine learning algorithms. They also reduce the in-out class variability compared to that of the raw pixels \citep{Lillywhite2013,Viola2001}. Haar features are mainly rectangles divided into black and white regions and the value of this feature is calculated by subtracting the sum of pixels in the white region from the sum of those in the black region \citep{Viola2001}. For each image, variations of each of the four Haar feature types are computed in all possible sizes and all possible locations, which provides a huge set of features. 

The authors chose Adaboost as a method to obtain their strong classifier. Adaboost was proposed by \citet{Freund1995}, it has the power to search through the features and select those of good performance then combine them to create a strong classifier. The general idea of the algorithm works as follows:
For a number of iterations T:
\begin{itemize}
	\item 	Pass through the set of all possible features and calculate the error of each one on the given images.
	\item 	Choose the best feature (the one with the lowest error) as the first weak classifier.
	\item Update the sample images corresponding weights, by putting more weights on the wrongly classified images.
	\item It then goes through the next iteration, until it finds the set of best features, to be used in classification.
\end{itemize}

One of the important contributions of  \citet{Viola2001} work is the cascade classifier which increased the accuracy while radically reducing the time consumed in detection. The cascade classifier is a stage classifier where the thresholds vary. The first stages has a low threshold, thus detecting all the true positives while eliminating the strong negatives, before more complex classifiers are used to achieve less false positives. Although their final detector has performed well in terms of accuracy, the training process is time consuming as Adaboost passes by the set of all possible features multiple times in training each stage of the classifier. In addition to that, training the stage classifier becomes slower in the last stages since the images become harder to classify. This is due to the use of images that were classified as false positives by the previous stages, therefore the last stages of the classifier typically need more features to satisfy the false alarm and the true positive rates desired. This means that the final stages has to go through the set of all possible features more times than in the first stages which puts an overhead on the training time.

Some of the researchers used the Viola-Jones algorithm as a base for their research to provide a more powerful detector.  \citet{Lienhart2002} proposed the increase of the Haar features used. Instead of using only the four basic types of Haar features, they increased them to include the 45 degree rotation of the original ones, consequently, the total number of feature types amounted to 14. The use of more features resulted in better accuracy, yet it substantially increased the number of generated features per image. A larger feature set means that more time will be taken by the training as the Adaboost operates in a brute force manner.

The previous work was concerned with enhancing the accuracy or speed of detection regardless of the overhead posed on the training time. This work aims to examine the effects on increasing the speed of training using GA and how this might affect the accuracy.

\section{\uppercase{PROPOSED METHOD}}
\label{sec:PROPOSED METHOD}

\noindent The proposed method (Named: GAdaBoost) applies GA to select a set of features, to have Adaboost choose from, instead of going through the set of all possible features. The original Adaboost algorithm was proposed by Freund and Schapire (1995) the generalized version works as follows: for the training of each stage in the stage classifier, the algorithm passes through the set of all possible features and calculates the error of each feature on each given image. After that, it chooses the best feature (the one with the lowest error, i.e best classifies the image correctly) as the first weak classifier. It then updates the sample images and their corresponding weights, by putting more weights on the wrongly classified images. The procedure is repeated until the set of chosen features reaches a preset false alarm, and hit rate set for classification.

\begin{figure}[h!]
	\centering
	\includegraphics[width=7cm]{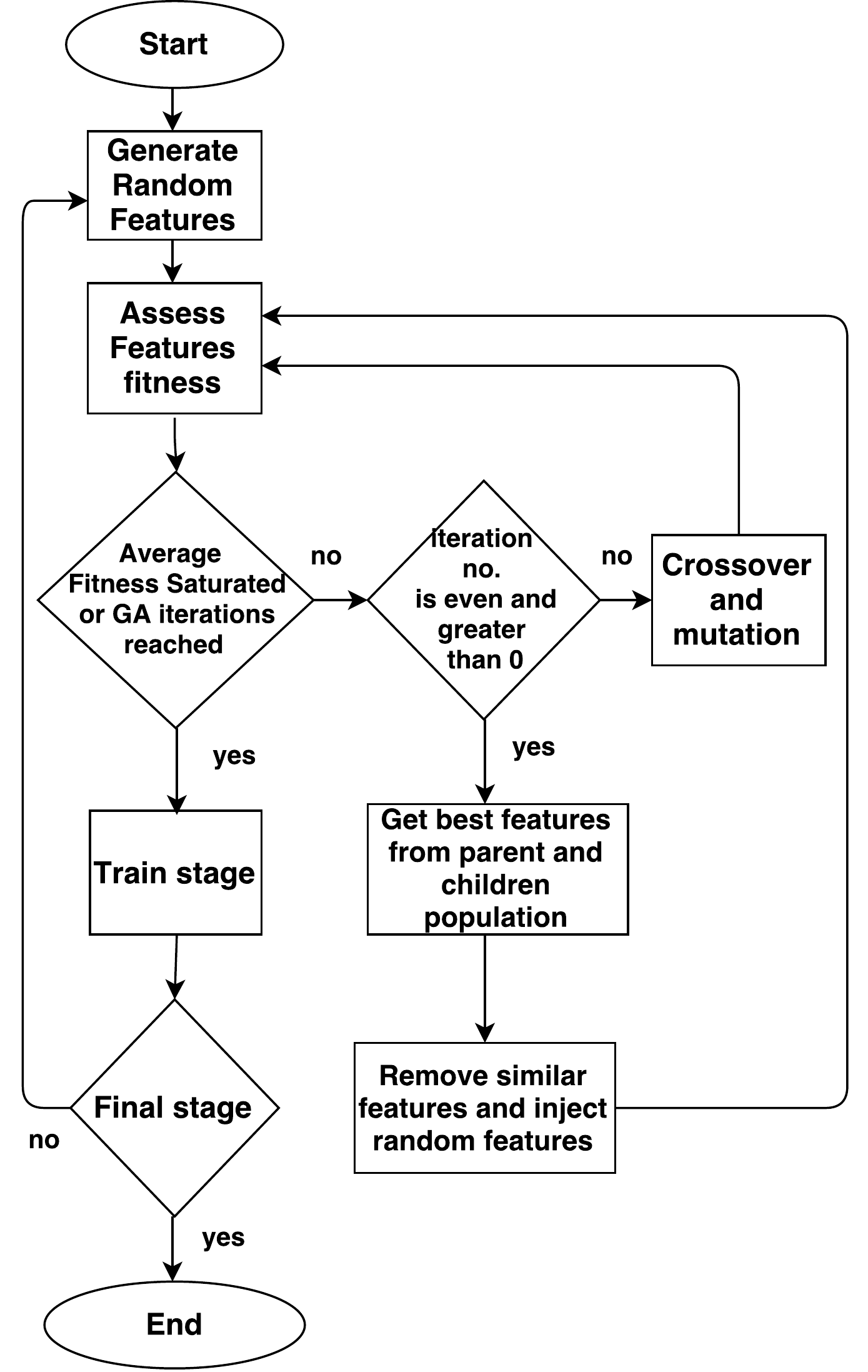}
	\caption{GAdaBoost diagram}
	\label{fig:AlgorithmFlowchartcameraReady}
\end{figure}

In order to  integrate the use of GA, OpenCV$'$s \citep{itseez2015opencv}  implementation on the Viola Jones algorithm has been modified.
Incorporating the use of GA will increase the training speed by avoiding the error calculation of the set of all possible features, and only providing the Adaboost algorithm with a representative set of features, that have been chosen based on their classification power.  This set of representative candidate features is to be prepared by the GA before the training of each stage in the final classifier. For example if the final classifier is to have 10 stages the added GA technique is to be repeated 10 times. The stage training utilizes Adaboost technique to choose multiple weak classifiers from the mentioned representative set, in order to reach the desired false alarm and hit rate preset for the stage. Figure \ref{fig:AlgorithmFlowchartcameraReady} shows a block diagram that explains the proposed GAdaBoost technique.

On the first iteration the GAdaboost chooses a preset number of features randomly to create the first generation of the given population size. Those randomly chosen features are marked so that they are not to be used again when more random features are to be generated. This is done to explore more of the set of all possible features. In order to assess the predictive power of these features, they are passed to a learning algorithm. The way this has been implemented is by creating a temporary (dummy) stage where the features are trained in the same way the original stage training works, i.e the dummy stage is an Adaboost training algorithm. The number of weak classifiers chosen by the Adaboost algorithm in the dummy stage  is a variable that is preset. The Adaboost algorithm associates the features with scores that are a representation of their predictive power. After that the best features are then selected and have mutation and crossover processes preformed on them to get the next generation of an even better preforming set of features. The new generation is then passed by a dummy stage for scoring. The process is repeated until the average fitness of the population saturates or a predefined number of iterations is reached.
As a form of exploring more of the set of all possible features, for each iteration with an even number (2nd, 4th, etc. generations) that is greater than zero, the best set of parents and their children produced are chosen. Then a spatial comparison is formed to remove the redundant features, and random features are inserted instead to complete the population size. The spatial comparison is done using the pasacal criterion where two features are considered of spatial similarity if the ratio of the intersection of the two features over the union of the two is greater than 0.4. This method is described in more details in section \ref{subsec:Training Speed up vs. Accuracy}. The use of only even iterations entails that the spacial comparison is done on half the number of iterations (eg for 50 iterations, the spatial comparison is done 25 times). The final set of features obtained by the GA is passed through a real stage where  the weak classifiers selected by this stage are to be used in the resultant final classifier. The afore mentioned technique ensures that the Adaboost algorithm will only evaluate the population size chosen instead of going through the whole set  of features when selecting the weak classifiers of the resultant final stage classifier.

\begin{figure}[!h]
	\centering
	\includegraphics[width=7cm]{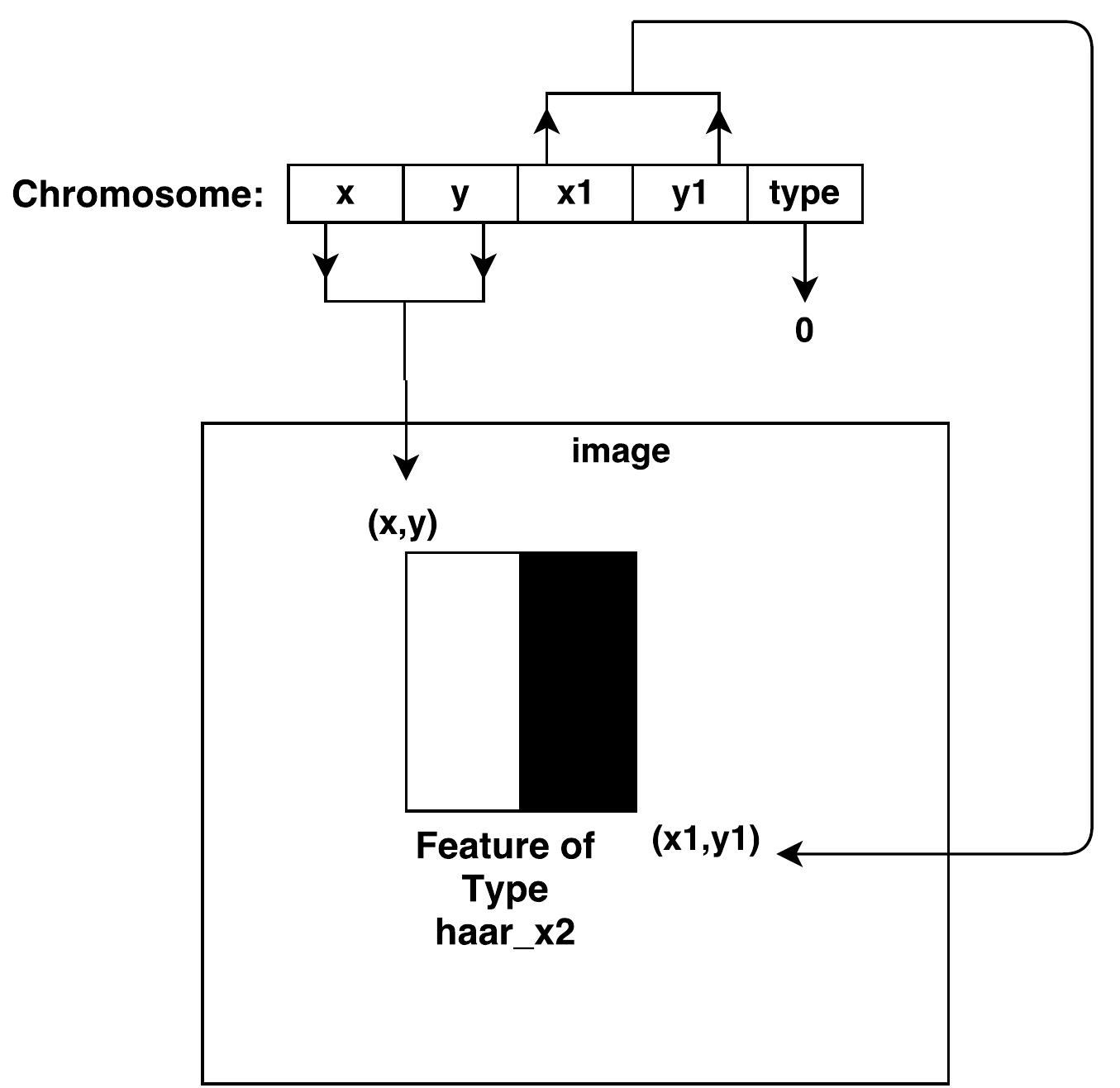}
	\caption{Chromosome to feature mapping}
	\label{fig:Chromosomerep}
\end{figure}

As for the Genetic Algorithm parameters, each chromosome represents one Haar feature. The values of the chromosome are x, y, x1, y1, type, where x, y are the upper left coordinates of the feature rectangle, and  x1, y1 are the lower right corner. The type is a value from 0-4 where each number represents one of the Haar Feature types used for upright frontal faces detection. 0,1,2,3,4 represent the Haar types of haar\_x2, haar\_y2, haar\_x3, haar\_y3, and  haar\_x2y2 respectively. Figure 2 explains the mapping of a feature of type haar\_x2 to a chromosome in a given image. As shown the chromosome carries decoded information about the type of the feature and its orientation in a given image, the way the chromosome is represented facilitates the mutation and the crossover processes which provides new features. The fitness function is a measure of how well this features splits between the negative and the positive images, or in other words predictive power of this feature in classifying the images correctly. The OpenCV implementation uses decision stumps as weak classifiers, these decision stumps are Classification and Regression trees (CART). In CART the regression tree’s best split quality is calculated by the minimization of Equation 1. 

\begin{equation}\label{eq1}
\sum_{i=1}^{n}(TR_i-PR_i)^2
\end{equation}

Where TR is the ground truth of the image, PR is the predicted responses by the decision stump and n is the number of sample images. Yet, for simplicity the OpenCV traincascade developer mentioned that in implementation the minimization criteria is reduced to equivalent simpler maximization ones \citep{dimashova2012}. The fitness of the feature used is the split quality measure provided by OpenCV$'$s CVDTree class. Thus, in the implementation the best feature is the one the largest quality.
The candidates to be used for crossover and mutation are selected using a Roulette Wheel selection method, so that those features with higher fitness are more likely to be selected and have the mutation and the crossover applied on them. A simple one-point crossover is utilized at the lower right corner of the two candidate features.
In order to reduce the time taken by validation of the correctness of the feature, the mutation is designed to assign the type to the feature according to how suitable this type is, given the coordinates of this feature.

\section{\uppercase{EXPERIMENTS AND RESULTS}}
\label{sec:EXPERIMENTS AND RESULTS}

\noindent In this section we discuss our experimental work, its setup, and the acquired results. It is divided into 3 subsections, each one describes an experiment setup and discusses the results acquired. The first section shows the fitness of the individual and the average fitness of the population. The second one shows the effects of varying the population size on the training time. The final one shows the effect of varying the number of iterations of GAdaBoost and comparing it against the original brute force algorithm with respect to the training time and accuracy. The testing was performed on both the FDDB dataset, and Caltech 10,000 Web Faces dataset. All the training occurred on the same computer with an Intel Core i7-4510U @ 2.00GHz processor and 8 GB RAM.

The positive images used for training are the images provided by OpenCV for the upright frontal faces training. The negative images were picked randomly from the dataset of 101 objects developed at Caltech \citep{LiFergusPerona04}. All the experiments use the same settings of a 17 stage cascade classifier with 500 positive images and 500 negative images per stage, hit rate of 0.9 and a false alarm rate of 0.5 per stage.  Each dummy stage is trained for 3 weak classifiers and the sample image weights are carried on between the dummy stages, with the exception of experiments in section 4.1 where each dummy stage has been trained for only one weak classifier, with no carrying on of the image weights between dummy stages, and the check on the special proximity and its removal wasn$'$t utilized.
\subsection{Individual and Population Fitness}

\noindent In this Experiment the GAdaBoost discussed in the proposed method (section 3) has been used to train a cascade classifier with the settings mentioned in the beginning of this section. A population size of a 1000 and 50 iterations were the parameters set for the GAdaBoost. Figure 3 shows the progress of the best individual, and the average fitness of the population. They are shown over the course of the 50 iterations of the GA preformed before the 17th stage.

\begin{figure}[!h]
	\vspace{-0.2cm}
	\centering
	%   {\epsfig{file = SCITEPRESS.eps, width = 5.5cm}}
	\includegraphics[width=7cm]{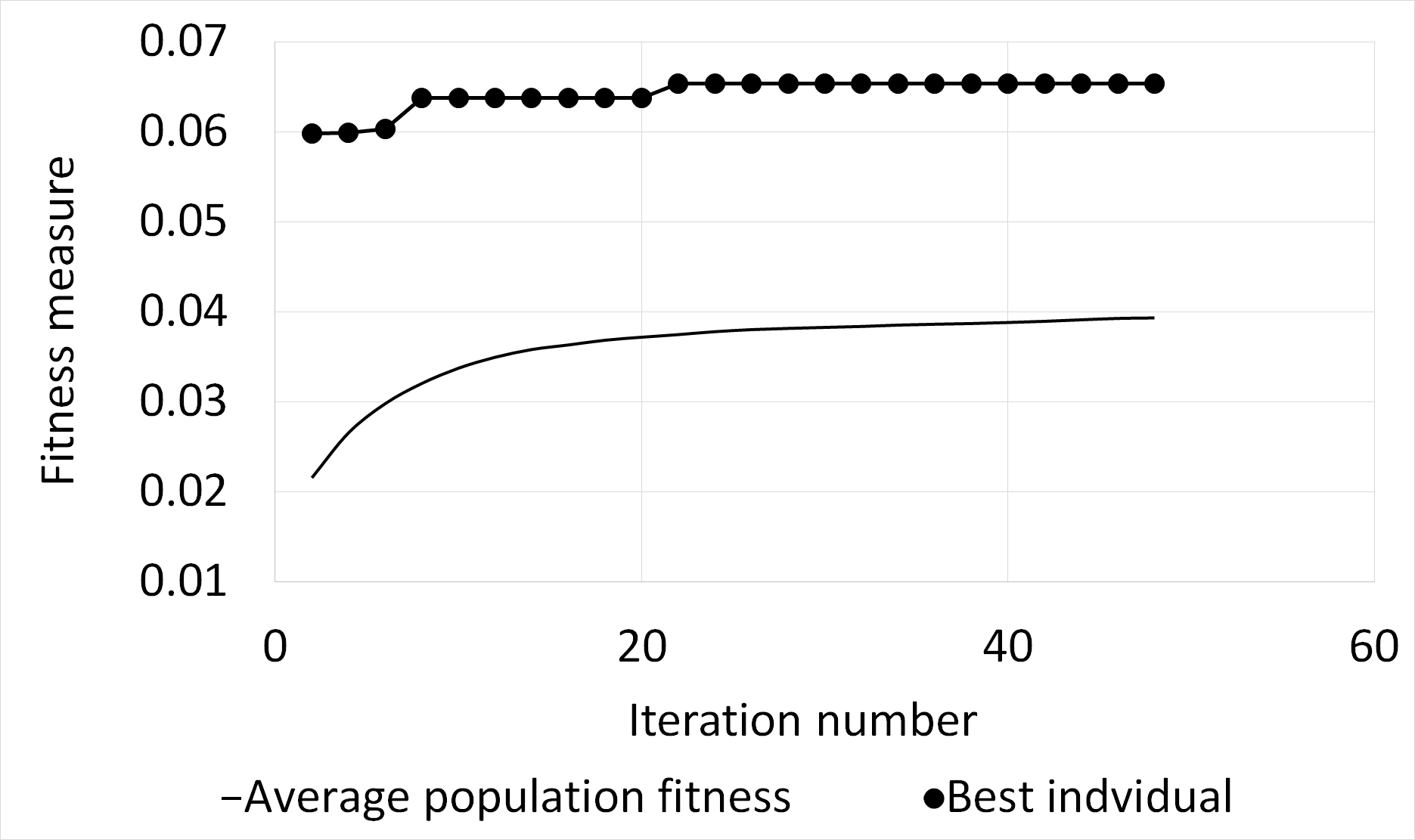}
	\caption{Best individual fitness and average population fitness over 50 iterations.}
	\label{fig:bestindvaveragefitness}
	\vspace{-0.1cm}
\end{figure}

As shown from Figure 3, the average population fitness increases fast in the first iterations then it starts to saturate, while the fitness trend of best individual is to increase then saturate and so on.

\subsection{Population Size vs Training Time}
\noindent In this Experiment 20 iterations is set for GAdaboost. Each classifier has been trained multiple times and the average time taken has been calculated. Figure 4 shows the effect of varying the population size on the training time.

From Figure 4 it is clear that the training time increases as the population size increases. This can be attributed to the fact that less mutations and crossovers are done with a smaller population size, and that the GA provides the Adaboost with a smaller feature set to go through in a brute force manner.

\begin{figure}[!h]
	%\vspace{-0.2cm}
	\centering
	%   {\epsfig{file = SCITEPRESS.eps, width = 5.5cm}}
	\includegraphics[width =7cm]{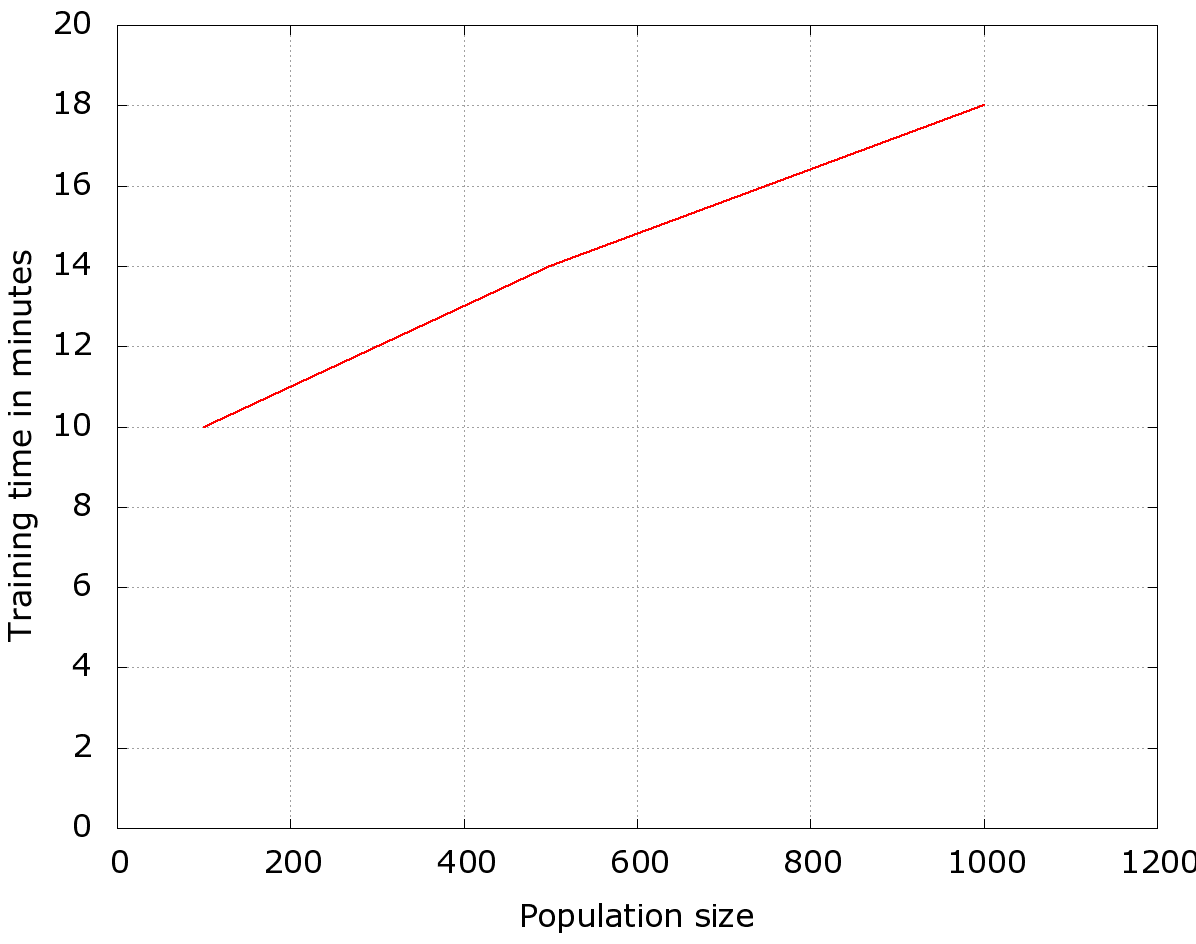}
	\caption{Population size vs training time.}
	\label{fig:timevspopsize}
\end{figure}
\subsection{Training Speed up vs. Accuracy}
\label{subsec:Training Speed up vs. Accuracy}

\noindent As a baseline a cascade classifier has been trained using OpenCV’s traincascade method using the same settings. This classifier has been used as a baseline to compare our algorithm with. We performed our experiment on both the Face Detection Dataset (FDDB)\citep{fddbTech}, and Caltech 10,000 Web Faces dataset \citep{Angelova2005}. Face detection datasets were chosen since Haar features were used originally to detect upright frontal faces. The FDDB dataset is a benchmark dataset designed for studying the unconstrained face detection problem. This dataset has been used in many studies and is considered one of the difficult datasets, due to occlusions, out of focus faces and difficult poses \citep{fddbTech}. This dataset contains annotations for 5171 faces in a set of 2845 images. In this paper 290 images from this dataset has been used for experimentation. For the evaluation of our detector on this dataset, we used the evaluation code provided by the authors of this dataset.

Caltech Web Faces is a dataset of human faces collected from the web \citep{Angelova2005}. It is a challenging dataset since it contains difficult examples such as extreme face orientations, occlusion like hats and glasses and variable light conditions \citep{Angelova2005}. The Caltech Web Faces data set consists of 10,524 annotated faces where the coordinates of the mouth, eyes and nose are given. For our experiments we randomly extracted 500 images to test on. We used the given eye coordinates to set a bounding square by assuming that the eye distance is half the face width. For evaluation, we use the pascal criteria shown in Equation 2 \citep{Everingham2010}.

\begin{equation}\label{eq2}
 \dfrac{area(B_{gt}\cap B_{det})}{area(B_{gt}\cup B_{det})}>0.4
\end{equation}

Where $B_{gt}$ is the ground truth bounding box and  $B_{det}$ is the detected bounding box. Thus the ratio of the area of intersection between the two boxes to the area of their union has to exceed 0.4 in order for the detected box to be counted as a face.

To test our algorithm we examined two variations of the GAdaBoost cascade classifier where we vary the number of iterations. GAdaboost has been trained using 20 and 50 iterations. Every experiment has been run multiple times, and an average of the training time and the performance of all the runs has been calculated. The speed of training vs accuracy of the two GAdaboost variations have been compared against the baseline. 

The obtained results show that training GAdaBoost in the two experiments took significantly less time than training the baseline. Training our approach with 20 and 50 iterations took about 26.8\% and 44.7 \% percent of the time taken to train the baseline respectively. Figure 5 visually emphasizes the difference in training time taken by the baseline and both algorithms.

Figures 6 and 7 provide the Y error bar graphs, showing the maximum, minimum and average results, for all the runs of both the 20 and the 50 iterations GAdaBoost on FDDB dataset. It’s clear from the figures that the 50 iterations GAdaboost preforms better. It even has best case scenarios where the accuracy was almost the same as the baseline, for certain numbers of false positives.
\begin{figure}[!h]
	\vspace{-0.2cm}
	\centering
	%   {\epsfig{file = SCITEPRESS.eps, width = 5.5cm}}
	\includegraphics[width=7cm]{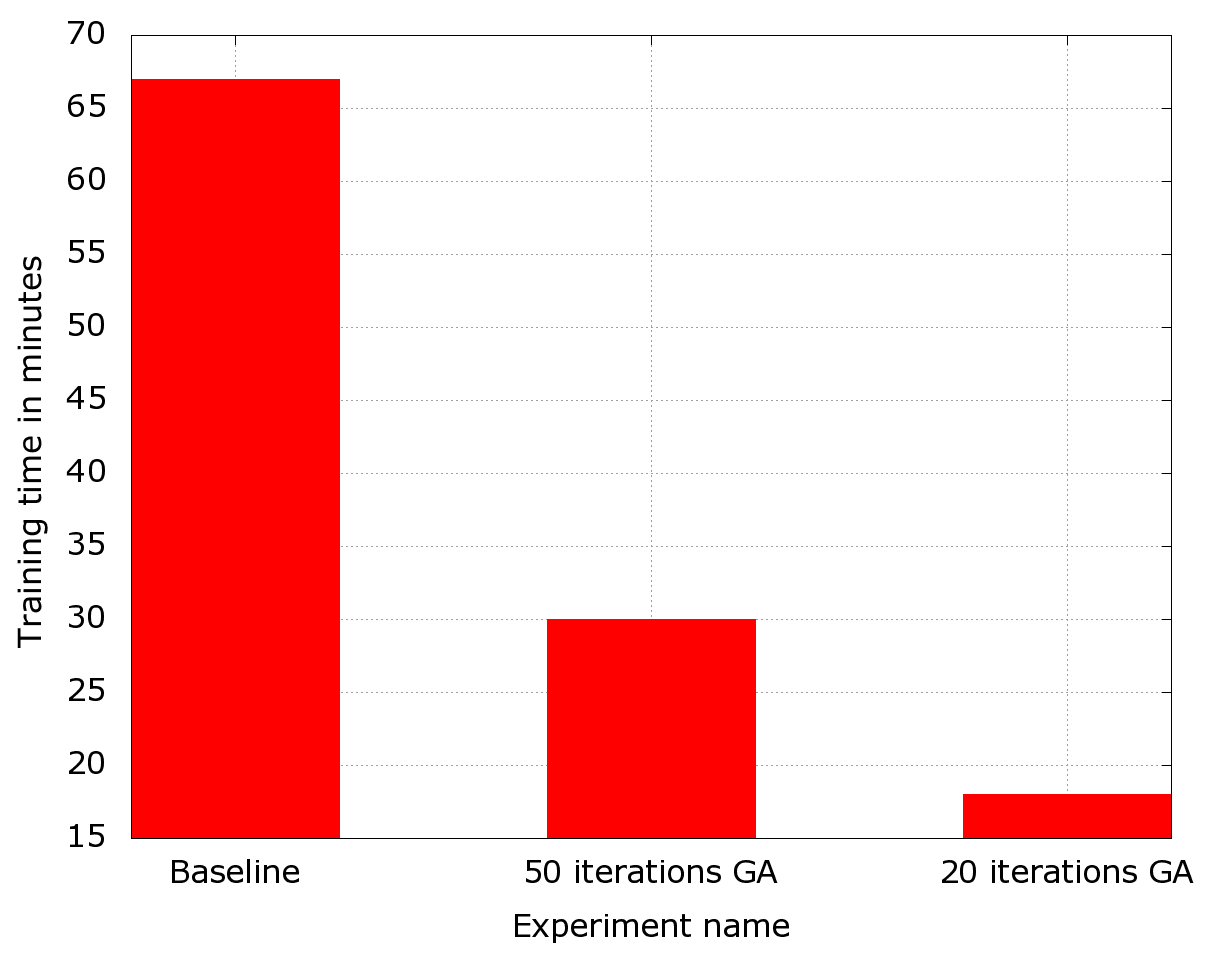}
	\caption{Training time in minutes of each of the experiments.}
	\label{fig:timecompare}
	\vspace{-0.1cm}
\end{figure}

\begin{figure}[!h]
	\vspace{-0.2cm}
	\centering
	%   {\epsfig{file = SCITEPRESS.eps, width = 5.5cm}}
	\includegraphics[width=7cm]{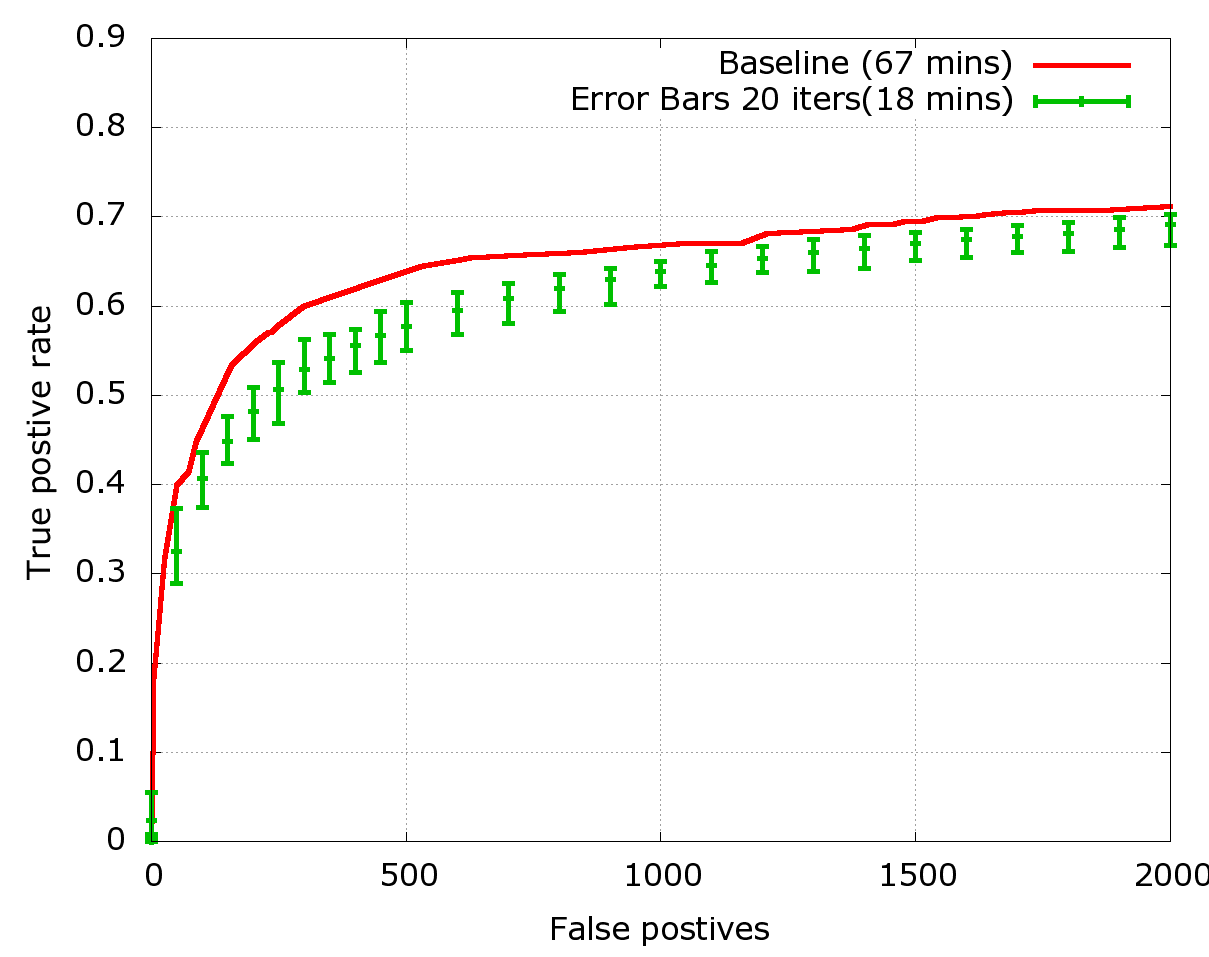}
	\caption{Y error bars for all the runs of the 20 iterations GAdaBoost on FDDB.}
	\label{fig:yerror20FDDB}
	\vspace{-0.1cm}
\end{figure}

\begin{figure}[!h]
	\vspace{-0.2cm}
	\centering
	%   {\epsfig{file = SCITEPRESS.eps, width = 5.5cm}}
	\includegraphics[width=7cm]{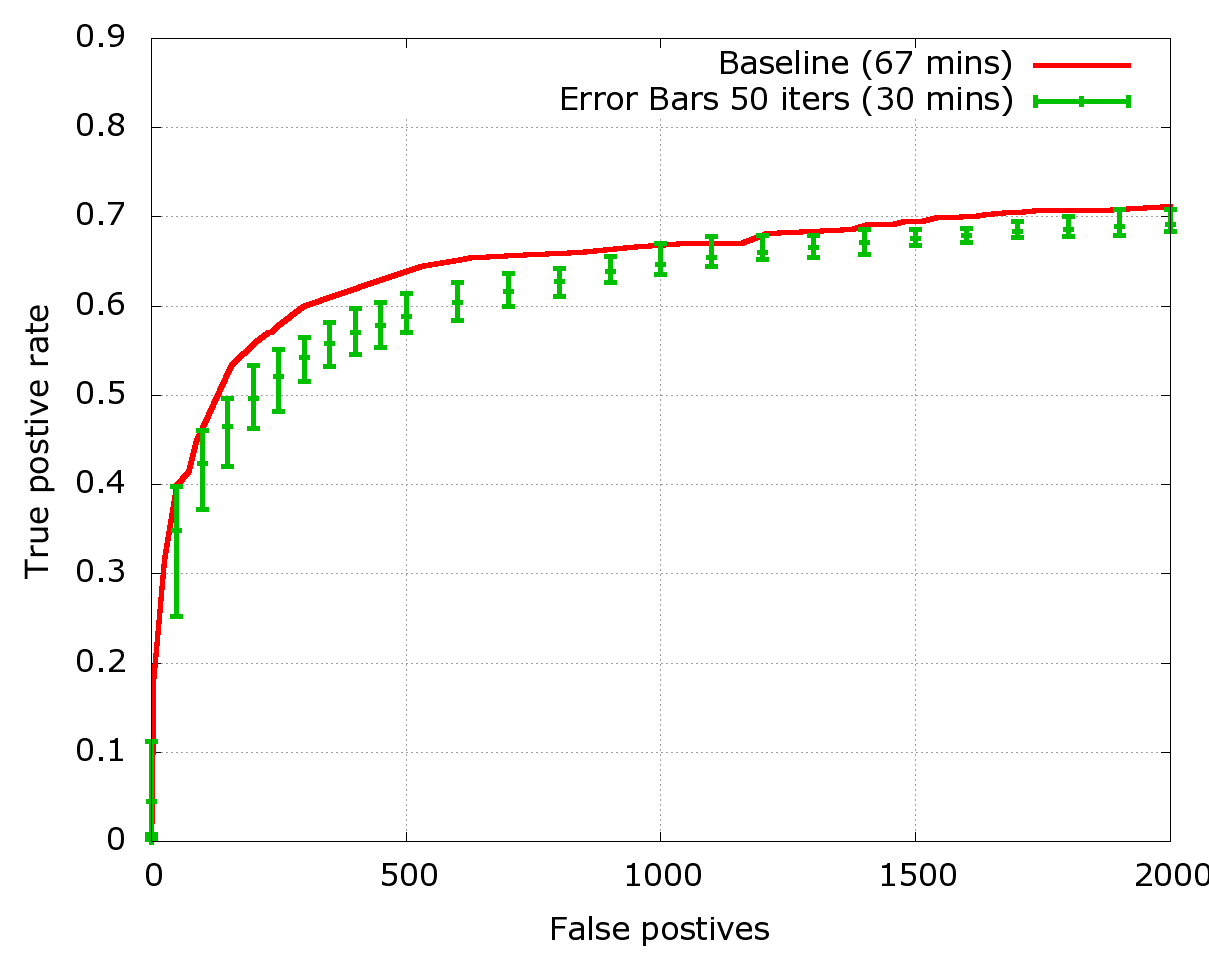}
	\caption{Y error bars for all the runs of the 50 iterations GAdaBoost on FDDB.}
	\label{yerror50FDDB}
	\vspace{-0.1cm}
\end{figure}

From both Figures 6 and 7, by examining the average point on the Y error bars it can be observed that at 500 false positives the baseline true positive rate is 64\% and the GAdaBoost 20 and 50 iterations achieved 58\% 59\% true positive rate respectively. While at 1000 false positives the baseline achieved 67\% true positive rate versus about 64\% and 65\% for the GA 20 and 50 iterations respectively.
Collectively from the provided figures, it can be noted that GAdaBoost with 50 iterations has performed slightly better than the GAdaBoost with 20 iterations. It can also be observed that at lower thresholds the GA provides closer true positive rates compared with the baseline, than it does at higher thresholds.

Figures 8 and 9 show the Y error bar graphs, showing the maximum, minimum and average results, for all the runs of both the 20 and the 50 iterations GAdaBoost on Caltech Web Faces dataset.

\begin{figure}[!h]
	\vspace{-0.2cm}
	\centering
	%   {\epsfig{file = SCITEPRESS.eps, width = 5.5cm}}
	\includegraphics[width=7cm]{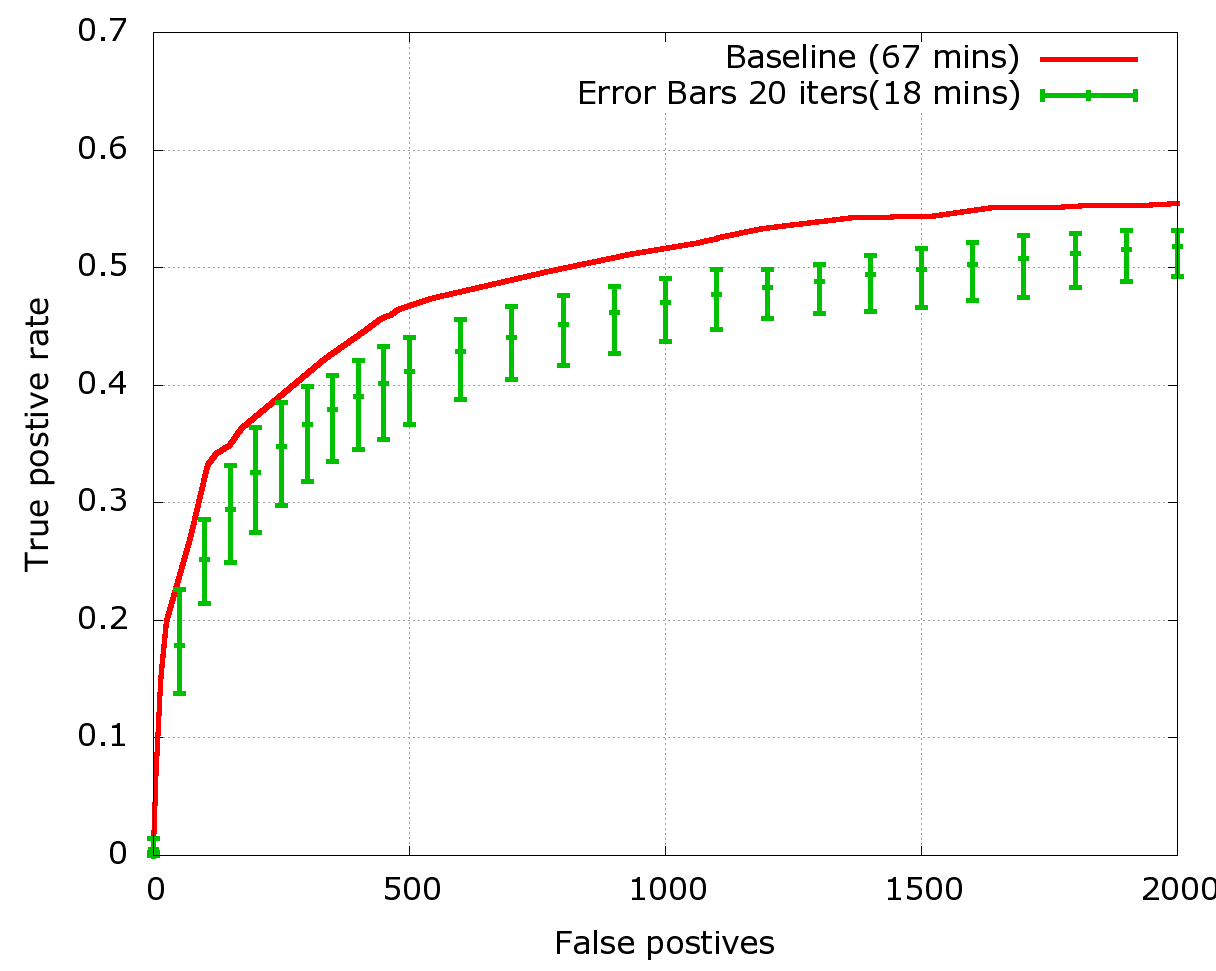}
	\caption{Y error bars for all the runs of the 20 iterations GAdaBoost on Caltech Web Faces.}
	\label{fig:yerror20caltech}
	\vspace{-0.1cm}
\end{figure}

\begin{figure}[!h]
	\vspace{-0.2cm}
	\centering
	%   {\epsfig{file = SCITEPRESS.eps, width = 5.5cm}}
	\includegraphics[width=7cm]{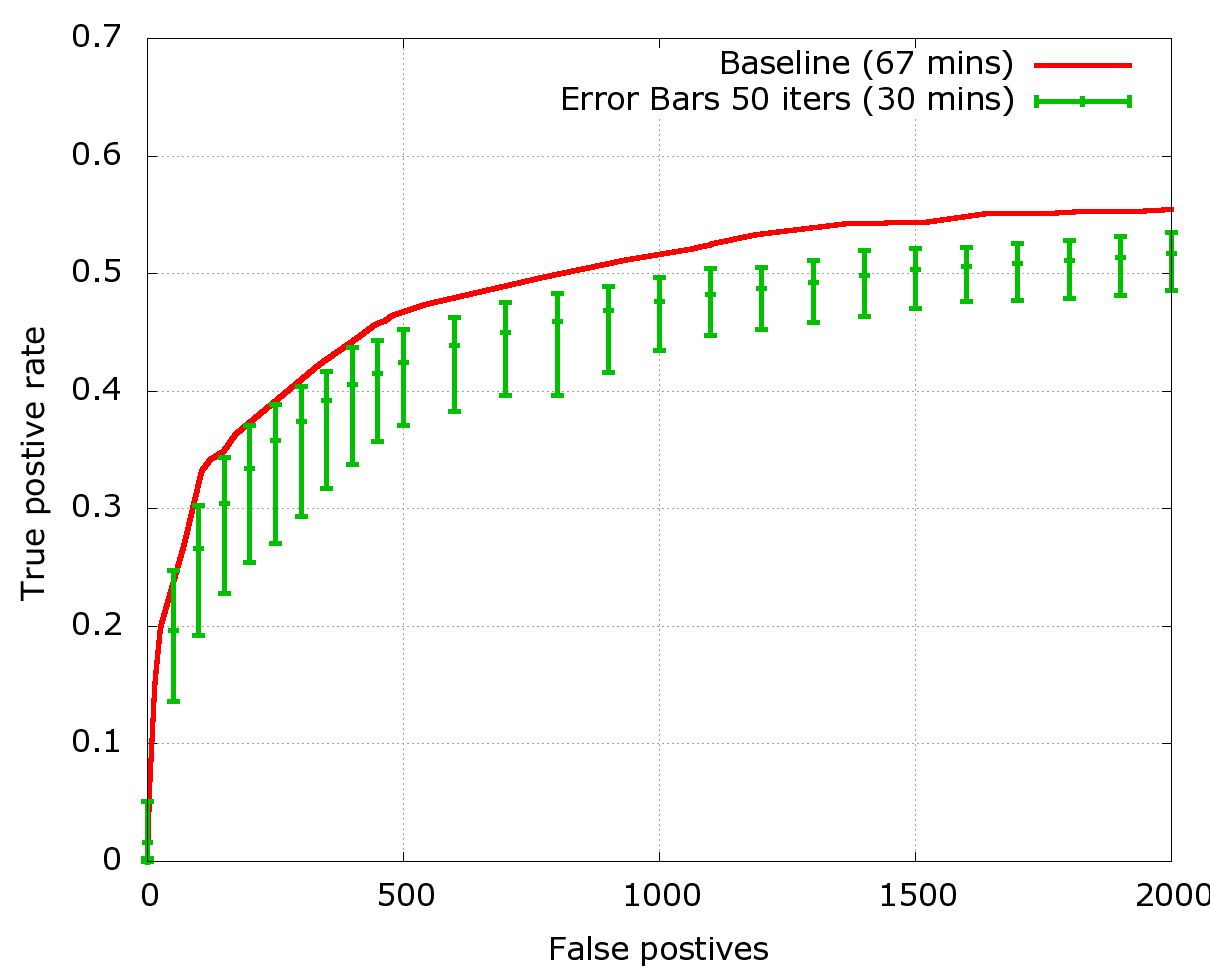}
	\caption{Y error bars for all the runs of the 50 iterations GAdaBoost on Caltech Web Faces.}
	\label{fig:yerror50caltech}
	\vspace{-0.1cm}
\end{figure}

From both Figures 8 and 9, by examining the average point on the Yerror bars we find that at 500 false positives the baseline true positive rate is 46 \% and the GAdaboost 20 and 50 iterations achieved 41\% , 43\% true positive rate respectively. While at 1000 false positives the baseline achieved 51\% true positive rate versus about 47\% and 48\% for the GAdaBoost 20 and 50 iterations respectively.
Collectively from the provided figures, it can be noted that GAdaboost with 50 iterations has performed slightly better than the GAdaboost with 20 iterations. 
The decrease in performance of both the baseline and GAdaboost can be attributed to the fact that Caltech Web Faces dataset includes occlusions and light variations as was mentioned at the beginning of this section.

\section{CONCLUSIONS}

\noindent We showed the effect of incorporating Genetic Algorithms with the Viola-Jones Rapid object detector on enhancing the training speed. Experiments to show the progression of the best individual and the average population fitness were provided. Other experiments showed the speed up that can be gained by the reduction of the population size. Also, two variations of the GAdaboost were examined, one with 20 iterations and the other with 50 iterations. Both experiments were run multiple times to observe the effect of the number of iterations on the performance using the FDDB and Caltech Web Faces dataset. We experienced that the training process became up to 3.7 times faster than the original algorithm with a mere decrease of 3\% to 4\% in accuracy. We noted that the 50 iterations performed better than the 20 iterations, and both had best case scenarios of almost reaching the baseline accuracy at some thresholds. 

The future extension of this contribution can be done by experimenting with more GAdaboost parameters by varying the iteration numbers, or finding better stopping criteria for the GA. The parallelizable nature of the GA can be utilized to gain an even faster training process.

\vfill
\bibliographystyle{apalike}
{\small
\bibliography{example}}

\begin{thebibliography}{}

\bibitem[Angelova et~al., 2005]{Angelova2005}
Angelova, A., Abu-Mostafa, Y., and Perona, P. (2005).
\newblock {Pruning training sets for learning of object categories}.
\newblock {\em Proceedings of the IEEE Computer Society Conference on Computer
  Vision and Pattern Recognition}, 1:494--501.

\bibitem[Chouaib et~al., 2008]{Chouaib2008}
Chouaib, H., Terrades, O.~R., Tabbone, S., Cloppet, F., and Vincent, N. (2008).
\newblock Feature selection combining genetic algorithm and adaboost
  classifiers.
\newblock In {\em Pattern Recognition, 2008. ICPR 2008. 19th International
  Conference on}, pages 1--4.

\bibitem[Dezhen and Kai, 2008]{Dezhen2008}
Dezhen, Z. and Kai, Y. (2008).
\newblock Genetic algorithm based optimization for adaboost.
\newblock In {\em Computer Science and Software Engineering, 2008 International
  Conference on}, volume~1, pages 1044--1047.

\bibitem[Dimashova, 2012]{dimashova2012}
Dimashova, M. (2012).
\newblock How is decision tree split quality computed.
\newblock
  http://answers.opencv.org/question/566/how-is-decision-tree-split-quality-computed/.

\bibitem[Everingham et~al., 2010]{Everingham2010}
Everingham, M., Gool, L.~V., Williams, C. K.~I., and Winn, J. (2010).
\newblock {The P ASCAL Visual Object Classes ( VOC ) Challenge}.
\newblock {\em International Journal}, pages 303--338.

\bibitem[Fei-Fei et~al., 2004]{LiFergusPerona04}
Fei-Fei, L., Fergus, R., and Perona, P. (2004).
\newblock {Learning Generative Visual Models From Few Training Examples: An
  Incremental Bayesian Approach Tested on 101 Object Categories}.
\newblock In {\em IEEE CVPR Workshop of Generative Model Based Vision (WGMBV)}.

\bibitem[Ferri et~al., 1994]{Ferri1994403}
Ferri, F., Pudil, P., Hatef, M., and Kittler, J. (1994).
\newblock Comparative study of techniques for large-scale feature selection*.
\newblock In GELSEMA, E.~S. and KANAL, L.~S., editors, {\em Pattern Recognition
  in Practice IVMultiple Paradigms, Comparative Studies and Hybrid Systems},
  volume~16 of {\em Machine Intelligence and Pattern Recognition}, pages 403 --
  413. North-Holland.

\bibitem[Freund and Schapire, 1995]{Freund1995}
Freund, Y. and Schapire, R. (1995).
\newblock {A desicion-theoretic generalization of on-line learning and an
  application to boosting}.
\newblock {\em Computational learning theory}, 55:119--139.

\bibitem[Itseez, 2015]{itseez2015opencv}
Itseez (2015).
\newblock {Open Source Computer Vision Library}.
\newblock https://github.com/itseez/opencv.

\bibitem[Jain and Learned-Miller, 2010]{fddbTech}
Jain, V. and Learned-Miller, E. (2010).
\newblock {FDDB: A Benchmark for Face Detection in Unconstrained Settings}.
\newblock Technical Report UM-CS-2010-009, University of Massachusetts,
  Amherst.

\bibitem[Lienhart and Maydt, 2002]{Lienhart2002}
Lienhart, R. and Maydt, J. (2002).
\newblock {An extended set of Haar-like features for rapid object detection}.
\newblock {\em Proceedings. International Conference on Image Processing},
  1:900--903.

\bibitem[Lillywhite et~al., 2013]{Lillywhite2013}
Lillywhite, K., Lee, D.~J., Tippetts, B., and Archibald, J. (2013).
\newblock {A feature construction method for general object recognition}.
\newblock {\em Pattern Recognition}, 46(12):3300--3314.

\bibitem[Sun et~al., 2004]{Sun2004}
Sun, Z., Bebis, G., and Miller, R. (2004).
\newblock {Object detection using feature subset selection}.
\newblock {\em Pattern Recognition}, 37(11):2165--2176.

\bibitem[Tabassum and Mathew, 2014]{Tabassum2014}
Tabassum, M. and Mathew, K. (2014).
\newblock {A Genetic Algorithm Analysis towards Optimization solutions}.
\newblock {\em International Journal of Digital Information and Wireless
  Communications (IJDIWC)}, 4(1):124--142.

\bibitem[Viola and Jones, 2001]{Viola2001}
Viola, P. and Jones, M. (2001).
\newblock {Rapid object detection using a boosted cascade of simple features}.
\newblock {\em Proceedings of the 2001 IEEE Computer Society Conference on
  Computer Vision and Pattern Recognition. CVPR 2001}, 1.

\bibitem[Xue et~al., 2015]{Xue2015}
Xue, B., Zhang, M., Member, S., and Browne, W.~N. (2015).
\newblock {A Survey on Evolutionary Computation Approaches to Feature
  Selection}.
\newblock 2007(September):1--20.

\end{thebibliography}

\vfill
\end{document}